\documentclass[10pt,twocolumn,letterpaper]{article}

\usepackage{iccv}
\usepackage{times}
\usepackage{epsfig}
\usepackage{graphicx}
\usepackage{amsmath}
\usepackage{amssymb}

\usepackage{graphicx}
\usepackage{amsmath}
\usepackage{amssymb}
\usepackage{booktabs}
\usepackage{multirow}

\usepackage{fp}
\usepackage{expl3}[2012-07-08]
\ExplSyntaxOn
\ExplSyntaxOff
\iccvfinalcopy

\usepackage{amsmath,amsfonts,bm}

\def\x{{x}}

\def\xi{{\x_i}}

\newcommand{\ignorethis}[1]{}

\def\eqref#1{equation~\ref{#1}}

\def\1{\bm{1}}

\def\vd{{\bm{d}}}

\def\vo{{\bm{o}}}
\def\vp{{\bm{p}}}

\def\vt{{\bm{t}}}

\def\vx{{\bm{x}}}

\DeclareMathAlphabet{\mathsfit}{\encodingdefault}{\sfdefault}{m}{sl}
\SetMathAlphabet{\mathsfit}{bold}{\encodingdefault}{\sfdefault}{bx}{n}

\newcommand{\ignore}[1]{}

\renewcommand*{\thefootnote}{\fnsymbol{footnote}}

\makeatletter
\DeclareRobustCommand\onedot{\futurelet\@let@token\@onedot}
\def\@onedot{\ifx\@let@token.\else.\null\fi\xspace}
\def\eg{e.g\onedot,\xspace} 

\def\ie{i.e\onedot,\xspace}

\makeatother

\newcommand{\final}[1]{\textcolor{black}{#1}}

\usepackage[pagebackref,breaklinks,colorlinks]{hyperref}

\usepackage[capitalize]{cleveref}
\crefname{section}{Sec.}{Secs.}
\Crefname{section}{Section}{Sections}
\Crefname{table}{Table}{Tables}
\crefname{table}{Tab.}{Tabs.}
\begin{document}
\title{AvatarCraft: Transforming Text into Neural Human Avatars with Parameterized Shape and Pose Control}
\author{
Ruixiang Jiang\\
The Hong Kong Polytechnic
University\\
{\tt\small rui-x.jiang@connect.polyu.hk}
\and
Can Wang \\
City University of Hong Kong \\
{\tt\small cwang355-c@my.cityu.edu.hk}
\and
Jingbo Zhang\\
City University of Hong Kong\\
{\tt\small jbzhang6-c@my.cityu.edu.hk}
\and
Menglei Chai\\
Google\\
{\tt\small cmlatsim@gmail.com}
\and
Mingming He \\
Netflix\\
{\tt\small hmm.lillian@gmail.com}
\and
Dongdong Chen \\
Microsoft Cloud AI\\
{\tt\small cddlyf@gmail.com}
\and
Jing Liao\footnotemark[1] \\
City University of Hong Kong\\
{\tt\small jingliao@cityu.edu.hk}
}

\maketitle
{
  \renewcommand{\thefootnote}%
    {\fnsymbol{footnote}}
  \footnotetext[1]{Corresponding Author.}
}

% \ificcvfinal\thispagestyle{empty}\fi

%%%%%%%%%%%%%%%%%%%%%%%%%%%%%%%%%%%%%%%%%%%%%%%%%%%%%%%%%%%%%%%%%%%%%%%%%%%%%%%%%%%%%%%%%%%%%%%%%%%
\begin{abstract}
Neural implicit fields are powerful for representing 3D scenes and generating high-quality novel views, but it remains challenging to use such implicit representations for creating a 3D human avatar with a specific identity and artistic style that can be easily animated. Our proposed method, AvatarCraft, addresses this challenge by using diffusion models to guide the learning of geometry and texture for a neural avatar based on a single text prompt. We carefully design the optimization framework of neural implicit fields, including a coarse-to-fine multi-bounding box training strategy, shape regularization, and diffusion-based constraints, to produce high-quality geometry and texture. Additionally, we make the human avatar animatable by deforming the neural implicit field with an explicit warping field that maps the target human mesh to a template human mesh, both represented using parametric human models. This simplifies animation and reshaping of the generated avatar by controlling pose and shape parameters. Extensive experiments on various text descriptions show that AvatarCraft is effective and robust in creating human avatars and rendering novel views, poses, and shapes. Our project page is: \url{https://avatar-craft.github.io/}.
\end{abstract}

% ========================================
% ========================================
\section{Introduction}
Creating human avatars is crucial for content generation in various immersive media, where users can alter the character to a specific identity, apply an artistic style, or animate with simple motion control.
While traditional manual authoring of digital characters often involves cumbersome and time-consuming efforts from skilled artists, the recent progress in human digitization has shown exciting potential towards more user-friendly solutions.
Nevertheless, avatar generation still faces a set of tough challenges. First of all, intuitive control is highly coveted for the system to understand specific user needs in the most natural form. Second, the generated avatars should be immediately ready for applications such as view synthesis, scene composition, and retargetable animation. Finally, the avatars should be of high quality, considering both the overall visual fidelity and preservation of target styles or identities in geometry and texture, especially when being manipulated or animated.

Significant efforts have been made in search of natural user controls for avatars. One representative stream of works~\cite{fivser2016stylit,bhatnagar2019multi,han2021exemplar,sang2022agileavatar,wang2021cross,texler2020interactive,texler2021faceblit} takes reference images to stylize an avatar.
Unfortunately, finding suitable references that perfectly match the desired shape and appearance is not always easy, substantially limiting real-world use.
On the other hand, text prompts are attracting more attention as a more natural control for generating high-quality 3D avatars, with the recent advances in large-scale vision-language models. In particular, text-to-3D avatar creation~\cite{hong2022avatarclip,youwang2022clip,michel2022text2mesh,wang2022nerf} is explored by leveraging the zero-shot generation ability of Contrastive Language-Image Pre-Training (CLIP)~\cite{radford2021learning}.
Following this trend, our work also tackles the problem of text-guided avatar creation. In particular, we aim at high-quality 3D avatar generation, which not only supports static view synthesis but also allows for controllable animation.

Text-driven avatar creation poses great challenges in producing high-quality geometry and texture 
while providing flexible animation capabilities. Existing methods~\cite{michel2022text2mesh, youwang2022clip, hong2022avatarclip} address these challenges by adopting cross-modal supervision to guide the generation and modeling avatars as explicit meshes to support skeleton-driven animation~\cite{hong2022avatarclip}.
However, despite the powerful representational capability of CLIP, due to the semantic structures and complex deformations of human bodies, these methods oftentimes struggle to produce detailed and consistent appearances for avatars~\cite{youwang2022clip}.
On a parallel thread, the pioneering text-conditional diffusion models~\cite{poole2022dreamfusion,lin2022magic3d} demonstrate stronger text-to-3D generation ability compared to CLIP. However, these approaches focus on general static objects rather than animatble human avatars.
To address these challenges, for the first time ever, we propose to tackle text-guided avatar creation leveraging a diffusion model for guidance, leading to improved results in terms of consistency and quality.

Additionally, instead of mesh-based representations~\cite{michel2022text2mesh,youwang2022clip}, we exploit neural implicit fields~\cite{muller2022instant,wang2021neus} to represent the avatar, which allow for volume rendering and generate high-quality novel views, making them especially advantageous for complex topology reconstructions and photo-realistic renderings. The implicit representation also makes it straightforward to composite the avatar with any implicit 3D scene while preserving realistic occlusions. To animate the avatar, we use SMPL to directly deform the neural implicit field, which enables a flexible way to animate and reshape the avatar by controlling the pose and shape parameters of SMPL, without requiring additional training.

In this paper, we propose \textit{AvatarCraft}, an approach for transforming text into neural human avatars, which uses diffusion models to stylize the geometry and texture, with shape and pose controlled by parametric human models. Our method starts with a bare neural human avatar as the template. Given a text prompt, we use diffusion models~\cite{rombach2022high} to guide the creation of a human avatar by updating the template with geometry and texture that are consistent with the text. However, directly applying diffusion models~\cite{rombach2022high} can lead to distorted geometry and textures as the diffusion loss~\cite{poole2022dreamfusion} is sensitive to the input resolution and biased towards geometry modeling. To address this issue, we design a novel coarse-to-fine multi-bounding-box training strategy, where the diffusion model guides the creation at multiple scales to improve the global style consistency while preserving fine details. We also introduce a shape regularization method to penalize the accumulated ray opacity of the avatar, to stabilize the optimization process. Regarding avatar animation, unlike skeleton-driven mesh animation in previous approaches~\cite{hong2022avatarclip}, our method defines an explicit warping field that maps the target human parameterization to the template human avatar and uses the warping field to deform the neural implicit field directly. Overall, our method enables parametrized shape and motion control of avatars and supports high-quality novel view synthesis, scene composition, and animation simultaneously. In summary, our contributions are as follows:

\begin{itemize}
\item We present a text-guided method for creating high-quality 3D human avatars that outperform previous approaches by using diffusion models as guidance.
\item Our method enables easy animation and reshaping of neural avatar radiance fields using only the pose and shape parameters of the SMPL model, without requiring any additional training.
\item We demonstrate the ability to composite our neural avatar radiance fields with real neural scenes for occlusion-aware novel view synthesis.
\end{itemize}

% ========================================
% ========================================
\section{Related Works}
\textbf{Neural Implicit Fields.} 
To represent 3D objects, previous approaches mainly utilize explicit representations like point clouds~\cite{cao2020psnet,lin2018learning}, voxels~\cite{guo2021volumetric,klehm2014property}, and meshes~\cite{kato2018neural,hollein2021stylemesh,han2021exemplar,zhang2020deep}. Due to the limited representative ability, these approaches can hardly synthesize high-fidelity novel views. Recently, led by the pioneering work of NeRF~\cite{mildenhall2020nerf}, the advances in neural implicit fields~\cite{mildenhall2020nerf,wang2021neus,yariv2021volume,muller2022instant,pumarola2021d,park2021nerfies} have sparked research into representing 3D objects via a continuous function, producing photo-realistic novel views by volume rendering. Specifically, NeuS~\cite{wang2021neus} and VolSDF~\cite{yariv2021volume} are proposed to improve NeRF~\cite{mildenhall2020nerf} by representing the geometry using a sign distance field (SDF) instead of occupancy.
Training such an implicit function takes a long time due to large amount of parameters in multi-fully connected layers. Instant-NGP~\cite{muller2022instant} reduces the training cost by imposing a smaller network without sacrificing quality. This network is augmented by a multiresolution hash table of learnable feature vectors, which significantly reduces the number of floating point and memory access operations, allowing for fast training of high-quality neural graphics primitives. Therefore, to improve the geometry and reduce the training cost, we combine NeuS and Instant-NGP as our basic architecture. Although the aforementioned methods improve convergence at the cost of a little precision, they can only model static scenes. Recent works~\cite{pumarola2021d,park2021nerfies,siarohin2023unsupervised,ren2021flow} extend the static neural implicit fields to dynamic ones with an implicit deformation network to warp sampled points along the ray to deform a template. Unlike them, instead of optimizing a deformation network to learn a warping field, we deform the neural implicit fields based on the parametric human model (i.e. SMPL). We define an explicit warping field by calculating a local transformation between the source SMPL mesh and the target one while aligning the source mesh to the neural implicit fields.

\textbf{Diffusion Models.} Recently, diffusion models have emerged as promising and widely-attracted image generators due to their impressive generative performance~\cite{ho2020denoising,nichol2021improved,nichol2021glide,saharia2022photorealistic,rombach2022high}. In addition to directly converting Gaussian noise into images with learned data distributions through iterative denoising, these models can also generate desired images conditioned on the guidance like class labels, text, and low-resolution images~\cite{dhariwal2021diffusion, rombach2022high, saharia2022palette,saharia2022image}. Text-driven diffusion models have demonstrated unprecedented capabilities in generating diverse high-quality semantically relevant images, such as GLIDE~\cite{nichol2021glide}, Imagen~\cite{saharia2022photorealistic}, and Stable Diffusion Model (SDM)~\cite{rombach2022high}. These models have been successfully applied in various domains, including image stylization, image editing, video generation, and 3D scene generation~\cite{huang2022diffstyler,zhang2022sine,hertz2022prompt, ho2022imagen,singer2022make,poole2022dreamfusion,lin2022magic3d}. Among them, DiffStyler~\cite{huang2022diffstyler} proposes a dual diffusion architecture for text-driven image stylization, which can generate text-dependent stylized images with the spatial structure consistent with the content image but cannot support novel view synthesis of stylized results due to the lack of 3D constraints. To bridge the gap between 2D images and 3D scenes, DreamFusion~\cite{poole2022dreamfusion} introduces a diffusion loss to enable 3D object generation from a pretrained 2D image-text diffusion model. Furthermore, Magic3D~\cite{lin2022magic3d} proposes a coarse-to-fine strategy for fine-grained 3D scene generation based on the pretrained diffusion model. Although DreamFusion and Magic3D enable the generation of view-consistent 3D objects from 2D prior models, 3D avatar creation remains a challenging task for them. We focus on investigating the text-guided generation of 3D human avatars with shape and pose flexibly controlled by the SMPL model.

\begin{figure*}[ht]
    \centering
    \includegraphics[width = 0.9\linewidth]{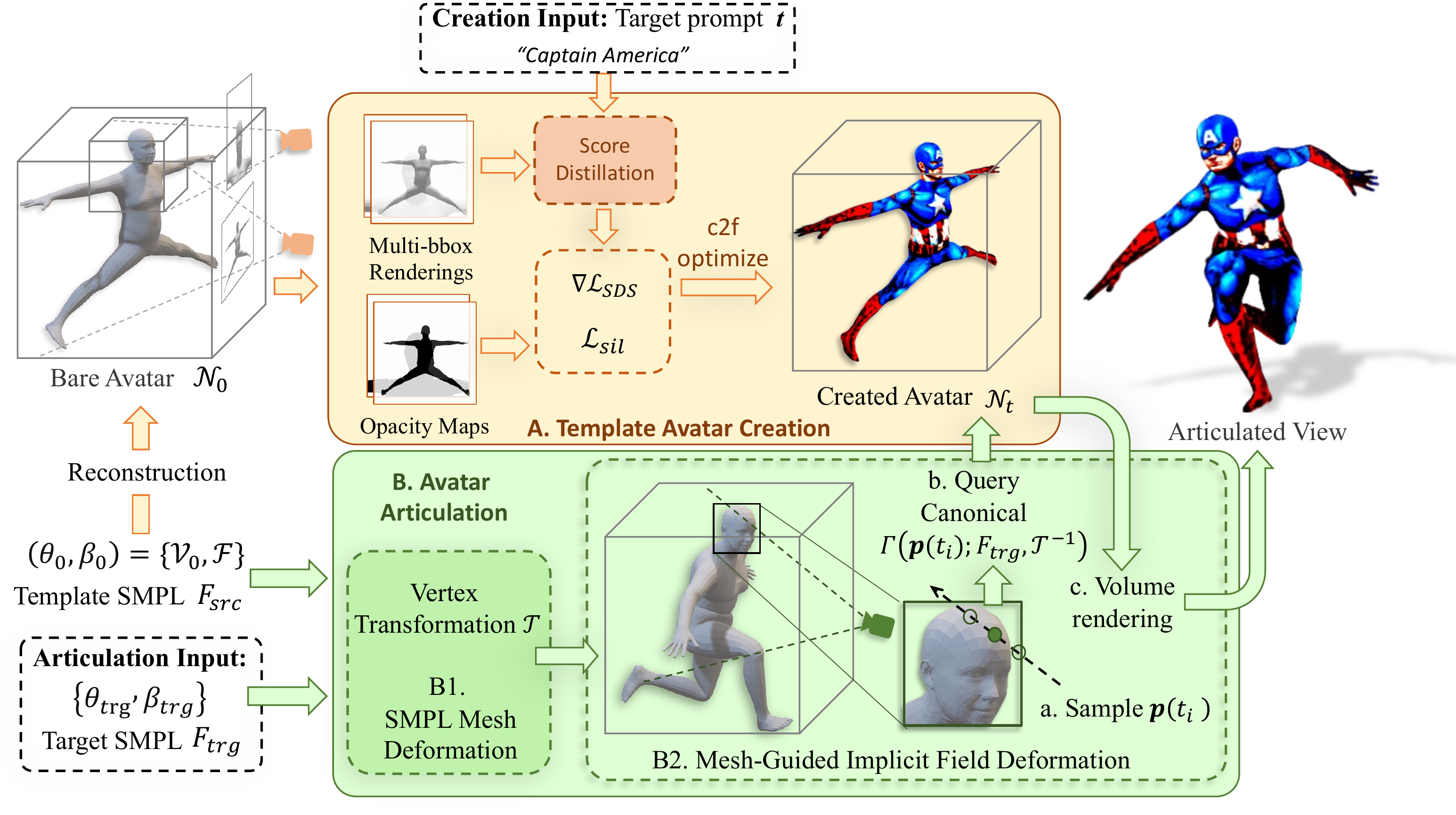}
    \caption{\textbf{Method Overview of AvatarCraft.} The proposed pipeline is divided into two stages. A) We utilize SDS loss and additional shape regularization to create the template of target avatar using our multiple bounding-box~(multi-bbox) and coarse-to-fine (c2f) training strategy. B1) We first use input SMPL parameters to calculate per-vertex rigid transformations. B2.a) Guided by $F_{trg}$, the camera emits rays, and points $\vp(t_i)$ on the ray are sampled. B2.b) For all sampled points, we find their corresponding points in the generated canonical space $\mathcal{N}_t$ based on inverse vertex transformation $\mathcal{T}^{-1}$ as well as SMPL mesh $F_{trg}$. B2.c) The color of the rays can be computed using the volumetric rendering equation.}
    % \vspace{-5pt}
    \label{fig: overview}
\end{figure*}

\textbf{Text-Guided 3D Avatar Creation.} Aside from creating 3D objects with reference images~\cite{chiang2022stylizing,nguyen2022snerf,huang2022stylizednerf,abdal20233davatargan}, the recent works propose to add detailed styles to a bare 3D body mesh given a text prompt as guidance. For example, Text2Mesh~\cite{michel2022text2mesh} and CLIP-Actor~\cite{youwang2022clip} utilize CLIP to guide the creation of a bare 3D mesh by learning a displacement map for geometry deformation and vertex colors for texture generation. Due to the limited representative ability of the mesh, such methods cannot generate detailed textures and render high-quality novel views. AvatarCLIP~\cite{hong2022avatarclip} and NeRF-Art~\cite{wang2022nerf} stylize a pre-trained neural implicit field guided by CLIP, producing photo-realistic renderings. AvatarCLIP first reconstructs a bare human avatar and then inpaints it using the classical CLIP similarity loss~\cite{patashnik2021styleclip}. It also designs a CLIP-guided method for reference-based motion synthesis to animate the stylized 3D avatar. NeRF-Art improves the stylization results by imposing a directional CLIP loss~\cite{gal2021stylegan} and a global-local contrastive loss. However, all these methods suffer from uneven and disordered textures generated by CLIP guidance. In addition, they lack the capability to animate a human avatar and render novel poses and views simultaneously, making them unfriendly to artists and designers. In contrast, we propose creating a 3D avatar under the guidance of diffusion models and elaborating a coarse-to-fine training strategy and shape regularization to produce visually pleasing results. Furthermore, we mitigate the animation issue by defining a local transformation between the template mesh and the target based on SMPL models, which enables easy control over the pose and shape of the stylized avatar.

\section{\final{Method}}

In this section, we introduce our method for creating and controlling neural implicit fields. First, we provide a preliminary explanation of our basic 3D representation in \S\ref{sec: prelim}. Next, we leverage diffusion models to guide the avatar generation direction \S\ref{sec: sds}. We also propose a shape regularization approach in \S\ref{sec: shape}, as well as coarse-to-fine and multi-bbox training strategies in \S\ref{sec: c2f}, to improve the generation performance. Finally, we animate and reshape the human avatar by defining an SMPL-guided deformation for the implicit fields in \S\ref{sec: warp}.

\subsection{\final{Neural Human Avatar Representation}}\label{sec: prelim}

NeuS~\cite{wang2021neus} improves upon the geometry of NeRF by replacing the occupancy representation in neural implicit fields with a SDF function. NeuS is composed of a geometry function $f(\vx):\mathbb{R}^3\mapsto\mathbb{R}^1$ and a color function $c(\vx):\mathbb{R}^3\mapsto\mathbb{R}^3$. The geometry function $f(\vx)$ takes 3D positions $\vx$ as input and regresses a zero-level set surface. 
On the other hand, the color function $c(\vx)$ takes $\vx$ and an optional view-direction $\vd$ as input, and outputs the radiance at that point. Similar to NeRF, NeuS also leverages volume rendering to achieve the pixel color of a ray $C(\vo,\vd)$, where $\vo$ and $\vd$ are the origin and direction of the ray, respectively. In addition, $n$ points are sampled $\{\vp(t)=\vo+t_i\vd\mid i=0,1,...,n\}$:
\begin{equation}
C(\vo,\vd)=\int^{\infty}_{0}{w(t)c(\vp(t),\vd)dt},\label{eq:render}
\end{equation}
where $w(t)$ is a weighting function:
\begin{equation}
w(t)=\frac{\phi_s(f(\vp(t)))}{\int_0^{\infty}{f(\vp(u))du}},
\end{equation}
where $\phi_s$ is the logistic density distribution, which allows for better geometry representation for human avatar modeling compared to NeRF.

Training deep coordinate-based networks can be challenging due to slow performance. However, a promising solution has been proposed by Instant-NGP~\cite{muller2022instant}, which introduces a multi-resolution hash encoding technique to alleviate this issue.
Specifically, it defines a multi-resolution voxel grid in space, as well as a table of learnable feature vectors corresponding to each voxel. To calculate the embedding at each voxel level for a query position $\vx$, the feature vector of that voxel is interpolated. The positional embedding of $\vx$ can be obtained by concatenating all the feature vectors. This streamlined embedding implementation in Instant-NGP has the potential to greatly enhance the reconstruction process while ensuring minimal loss in quality. Despite its strengths, Instant-NGP has not yet achieved the same level of geometry reconstruction accuracy as NeuS.

We combine NeuS and Instant-NGP as our neural implicit model of avatars, leading to an improvement in speed while maintaining satisfactory reconstruction quality.

\subsection{Diffusion-Guided Avatar Creation} \label{sec: sds}

We start by introducing diffusion-guided avatar creation from a text prompt. Let $\mathcal{N}_0=\{f(\vx),c(\vx)\}$ denote the bare avatar in canonical space, and $\vt$ be a text prompt. Our objective is to optimize \final{$\mathcal{N}_0$} such that the generated avatar $\mathcal{N}_\vt$ is amenable to $\vt$. 
\final{To achieve this,} we first build $\mathcal{N}_0$ \final{by reconstructing from} multi-view renderings of a bare SMPL mesh $\mathcal{M}_0=\{\theta_0,\beta_0\}=\{\mathcal{V}_0,\mathcal{F}\}$, where $\theta_0$ and $\beta_0$ are SMPL pose and shape parameters, and $\mathcal{V}_0$ and $\mathcal{F}$ represent vertices and faces, respectively.
After reconstructing this template, we leverage the recent Score Distillation Score (SDS) loss from DreamFusion~\cite{poole2022dreamfusion} to guide the creation process, which is defined as:
\begin{equation}
\nabla\mathcal{L}_{SDS}=\mathbb{E}_{m,\epsilon}\left[s(m)\left(\epsilon_\phi\left(z_m;m,\vt\right)-\epsilon\right)\frac{\partial z_m}{\partial x}\frac{\partial x}{\partial \theta}\right],
\end{equation}
where $\epsilon_\phi$ is the denoiser, $s(m)$ is a weighting function depending on the timestep $m$, $z_m$ is the noise, and $x$ is the latent code encoded from the 2D rendering of the avatar. \final{ By computing the SDS
loss, we enable the propagation of gradients from the diffusion model to update the neural radiance field.}

\subsection{Shape Regularization}\label{sec: shape}

\begin{figure}
    \centering
    \includegraphics[width=0.95\linewidth]{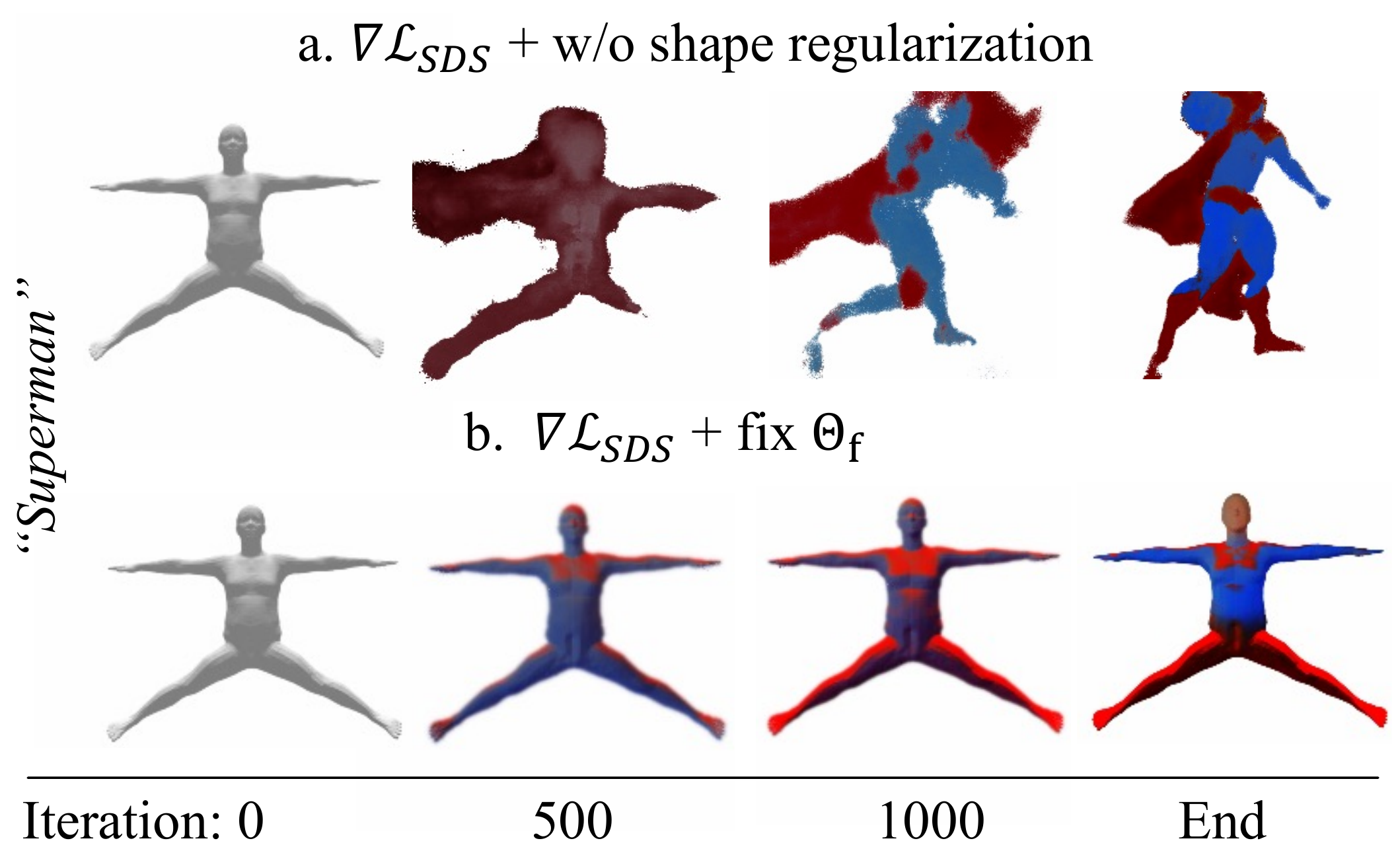}
    \caption{\textbf{Problems with Applying $\nabla\mathcal{L}_{SDS}$ for Avatar Generation}. We demonstrate the optimization progress of generation under two conditions: a) without any geometry constraint, and b) with the SDF parameters $\Theta_f$ fixed. Applying no constraint leads to adversarial results, while fixing the SDF parameters results in blurry texture. The prompt for both experiments is \textit{``Superman''}.}
    % \vspace{-5pt}
    \label{fig: geo_constraint} 
\end{figure}

Previous approaches~\cite{poole2022dreamfusion,lin2022magic3d, metzer2022latent} demonstrate the power of SDS loss in supervising 3D generation tasks. However, directly applying it to shift the shape $f(\vx)$ and color $c(\vx)$ of a human avatar to match the target style description $\vt$ can yield undesired results, as shown in Fig.~\ref{fig: geo_constraint}. The SDS loss is biased towards modeling geometry $f(\vx)$ in the early generation steps, which can result in the \final{catastrophic} destruction of the template prior $\mathcal{N}_0$ and an attempt to generate the human avatar from scratch instead. The high variance of the diffusion model in this generation process can be unstable, resulting in degenerate solutions such as empty volumes, or adversarial results such as flat geometry and multi-face issues~\cite{poole2022dreamfusion,lin2022magic3d}. Therefore, to stabilize our generation process, we propose to regularize the shape $f(\vx)$ as below.

We denote the \final{trainable parameters of the} geometry and color network in neural implicit fields as $\{\Theta_f,\Theta_c\}$. A naive solution to regularize the geometry is to freeze $\Theta_f$. However, as shown in Fig.~\ref{fig: geo_constraint}, this simple scheme leads to \final{blurry} and over-saturated results as it prevents the neural implicit fields from learning high-frequency details.
Therefore, we aim to optimize \final{both $\{\Theta_f,\Theta_c\}$}, while regularizing the \final{optimization of shape network}. Our key idea is that the object silhouette could serve as a proxy for its geometry, while allowing for non-convex details to be sculpted~\cite{laurentini1994visual}. Specifically, a pixel-level loss is introduced to regularize the ray opacity of the generated avatar:  
\begin{equation}
\mathcal{L}_{sil} =\frac{1}{HW}\sum_{H,W}|O(\vo,\vd;\mathcal{N}_0)-O(\vo,\vd;\mathcal{N}_\vt)|,
\end{equation}
where $O(\vo,\vd;\mathcal{N})$ denotes the accumulated point opacity along the ray direction $\vd$:
\begin{equation}
O(\vo,\vd;\mathcal{N})=\int^{\infty}_{0}{w(t)dt}.
\end{equation}

Finally, the overall loss used in the generation process is:
\begin{equation}
    \mathcal{L} = \nabla\mathcal{L}_{SDS} + \lambda_{sil}\mathcal{L}_{sil} +\lambda_{eik}\mathcal{L}_{eik},
\end{equation}
where $\mathcal{L}_{eik}$ is the Eikonal term~\cite{gropp2020implicit} to regularize the normal.

\begin{figure}[t]
    \centering
    \includegraphics[width=0.9\linewidth]{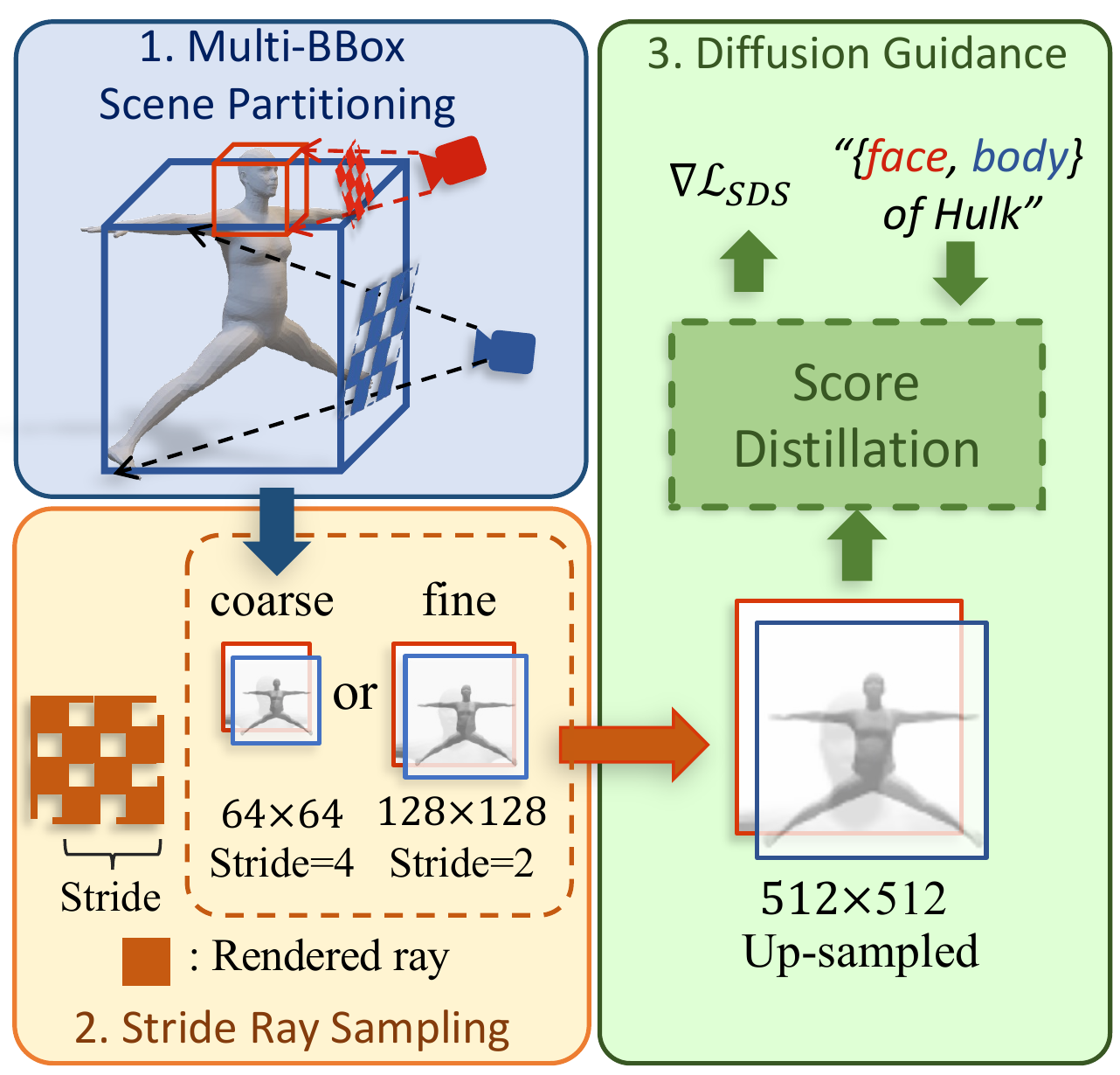}
    \caption{\textbf{Coarse-to-Fine and Multi-BBox Training.} 1) We partition the canonical space into two bounding boxes for sampling cameras. 2) we use stride ray sampling to render the avatar at different scales. 3) the rendered coarse or fine avatar is interpolated to fit stable diffusion input assumption for calculating the SDS loss.}
    % \vspace{-5pt}
    \label{fig: c2f}
\end{figure}

\subsection{Coarse-to-Fine and Multi-BBox Training}\label{sec: c2f}

A key to improving the fidelity of a generated avatar lies in allocating the correct texture to each part of the human body. Taking the style \textit{``Superman''} as an example, users would typically expect a large ``S'' on the chest and a red crotch to be the most significant details of Superman's costume. Subsequently, users might expect additional details in the face, belt, and shoes that resemble the identity of Superman. Visual features related to these attributes span different scales, and creating an avatar at one specific scale may result in the loss or misalignment of important details. Based on this observation, we propose a coarse-to-fine generation strategy, which aims to capture style details at different scales. 
Specifically, we adopt a two-stage generation scheme. Firstly, we stylize $\mathcal{N}_0$ in the rendering resolution of $64\times64$. Then, we fine-tune it by doubling the resolution while using the same set of losses to generate texture details. The lower resolution of the initial renderings serves as a natural band-limited filter that promotes the creation of high-level textures. As the resolution increases, the generation of fine-detailed textures becomes feasible.

In addition, human perception is particularly sensitive to artifacts and distortions in facial features. However, directly stylizing the human avatar radiance fields can result in a degradation of facial features. To address this limitation, we devise a solution to divide the scene into face and body bounding boxes according to the SMPL prior $\mathcal{M}_0$. Our approach involves dedicating the face box exclusively to rendering the head and neck, while the body box is used to render the entire avatar, including the head. Besides, we also augment the prompt $\vt$ according to the rendered box. This approach enhances the fidelity of facial features while maintaining a natural transition between the head and the body. By employing this coarse-to-fine and multi-bounding-box training strategy, we can improve the alignment of visual features across the entire avatar body.

\subsection{SMPL-Guided Avatar Articulation} \label{sec: warp}

Avatar $\mathcal{N}_\vt$ in the canonical space is intrinsically aligned with the underlying SMPL model $F_{src}=\{\theta_{0},\beta_{0}\}=\{\mathcal{V}_{0},\mathcal{F}\}$. Given the target model $F_{trg}=\{\theta_{trg},\beta_{trg}\}=\{\mathcal{V}_{trg},\mathcal{F}\}$, our method aims to deform the template avatar to render novel views with the target pose and shape. We demonstrate that our approach provides an intuitive way to control both the shape and the pose of the implicit avatar without the need for training, enabling various applications such as avatar shape customization and animation.

\textbf{SMPL Mesh Deformation.} We first establish the correspondence between $F_{src}$ and $F_{trg}$. Specifically, for each vertex, we use beta-blended shape~\cite{loper2015smpl} to compute the shape-related vertex deformation $\mathcal{T}^{\beta_{trg}}\in \operatorname{SE}(3)$ from $\{\theta_{0},\beta_{0}\}$ to $\{\theta_{0},\beta_{trg}\}$. Next, we calculate the shape-related vertex rigid transformations $\mathcal{T}^{\theta_{trg}}\in \operatorname{SE}(3)$ from $\{\theta_{0},\beta_{trg}\}$ to $\{\theta_{trg},\beta_{trg}\}$ using Linear Blended Skining (LBS) algorithm. Since SMPL mesh topology is agnostic to shape and pose by definition, we can obtain a bijective mapping of vertices between the two meshes as $\mathcal{T}=\mathcal{T}^{\theta_{trg}}\mathcal{T}^{\beta_{trg}}$.

\textbf{Mesh-Guided Implicit Field Deformation.} After obtaining the vertex mappings $\mathcal{T}$, we warp the positions of the sampled points to render novel views with the target shape and pose. We rewrite the volume rendering as:
\begin{equation}
C(\vo,\vd)=\int^{\infty}_{0}{\eta(\vp(t))\left[w(t)c(\Gamma(\vp(t);F_{trg},\mathcal{T}^{-1}))\right]dt},
\end{equation}
where $\Gamma(\vx;F_{trg},\mathcal{T}^{-1})$ is a transfer function that maps points $\vx$ from the observation space to the canonical space, guided by the target mesh and the inverse vertex  $\mathcal{T}^{-1}$~\cite{jiang2022neuman,yuan2022nerf}. To mitigate cloudy artifacts due to inaccurate warping, we adopt a mask function $\eta(\vp(t))$~\cite{chen2021animatable} to set the density of the points far from the target mesh surface to zero.

\begin{figure}
    \centering
    \includegraphics[width=0.95\linewidth]{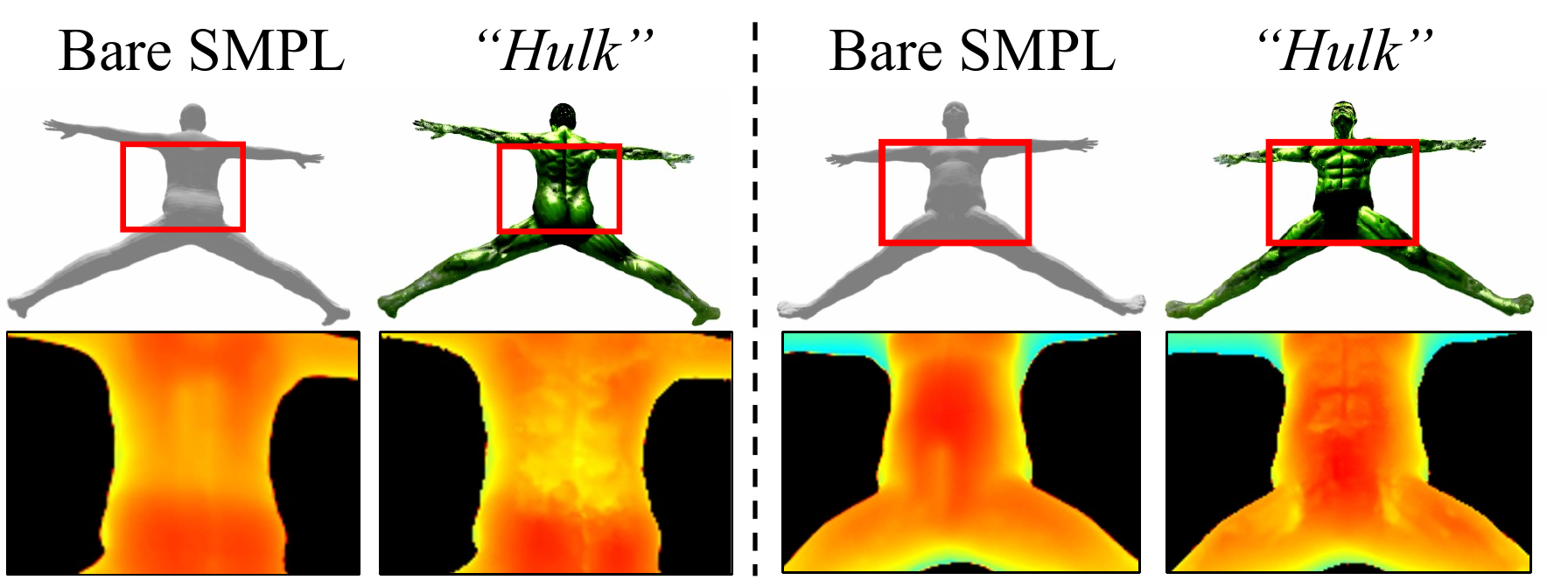}
    \caption{\textbf{Geometry Generation.} AvatarCraft could generate fine geometry detail on the avatar surface. We show the rendered depth map of bare SMPL and \textit{``hulk''}.}
    \label{fig:geo_edit}
\end{figure}

\begin{figure}
    \centering
    \includegraphics[width=0.95\linewidth]{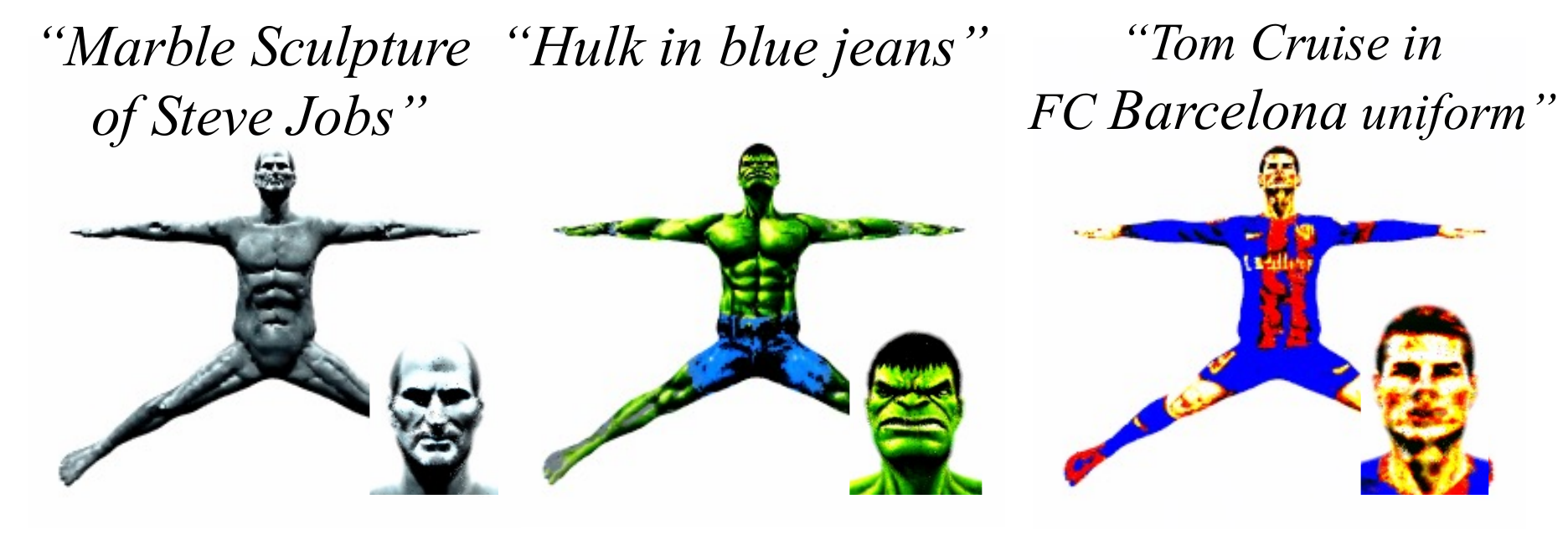}
    \caption{\textbf{Concept Mixing.} AvatarCraft could generate novel avatars by mixing different concepts together.}
    \label{fig:mix}
\end{figure}

\begin{figure}
    \centering
    \includegraphics[width=0.95\linewidth]{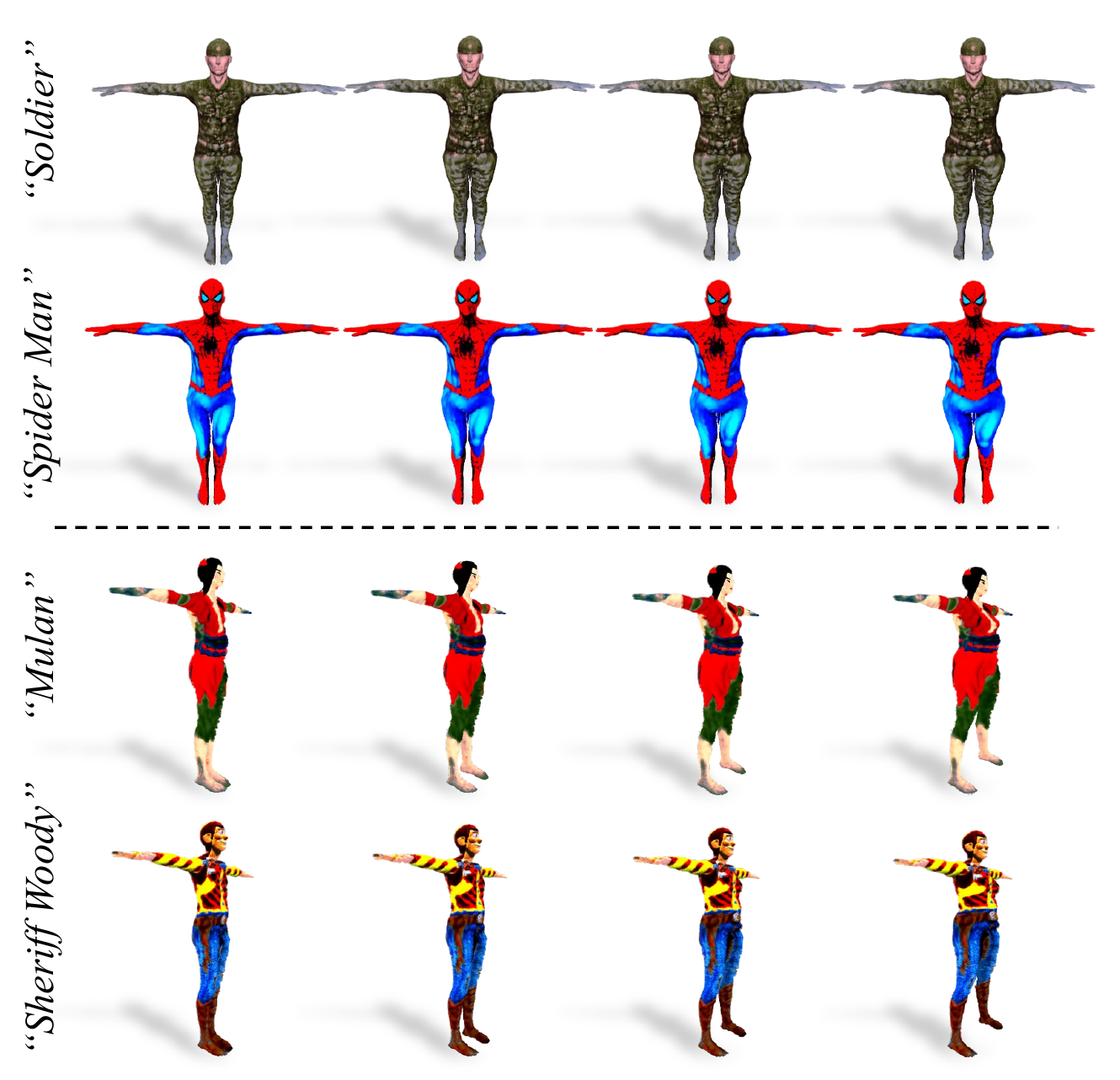}
    \caption{\textbf{Shape Interpolation.} AvatarCraft can reshape the generated avatar without the need for retraining. We demonstrate two different interpolation sequences of shape $\beta$ across four styles.}
    % \vspace{-5pt}
    \label{fig:shape}
\end{figure}

% ========================================
% ========================================

\section{Experiment Result and Analysis}

\begin{figure*}
    \centering
    \includegraphics[width=.9\linewidth]{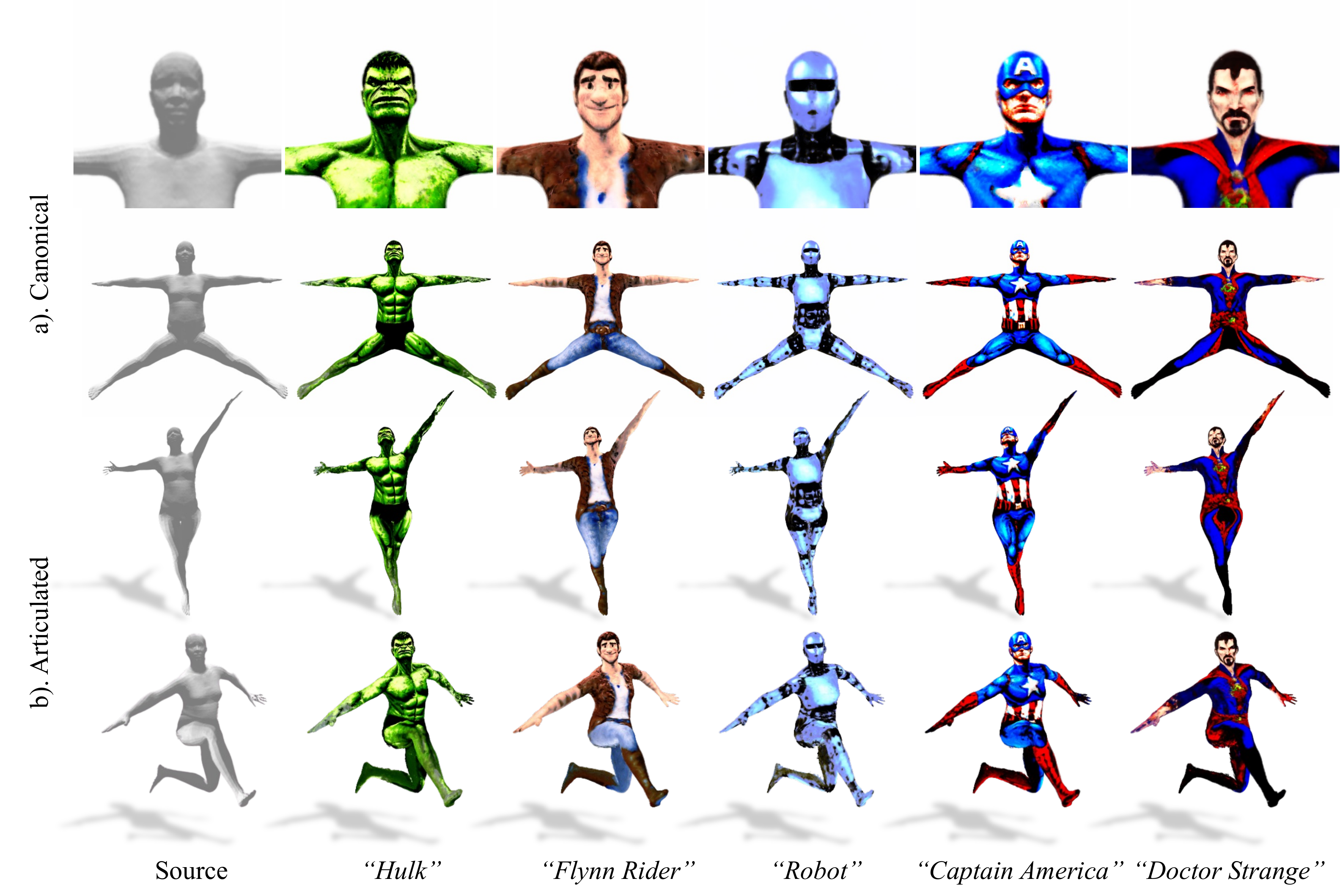}
    \caption{\textbf{Qualitative Result. }\final{a.) AvatarCraft generates intricate details that is consistent and aligned with input prompt. b). The generated avatar could be articulated by shape and pose parameters.}}
    \label{fig: result}
\end{figure*}

\subsection{Implementation Details}

In terms of network structure, we utilize the default setting for the hash-grid~\cite{zhao2022human}. Additionally, the SDF network has a depth of $2$, while the color network has a depth of $3$.
We adopt similar augmentation strategies to those used in recent text-to-3D generative models~\cite{hong2022avatarclip, poole2022dreamfusion,lee2022understanding,jain2022zero,lin2022magic3d}. Specifically, these include: (1) random camera extrinsic augmentation, (2) random background augmentation, and (3) view-dependent prompt augmentation. Further details about each of these augmentations can be found in the Appendix.

We reconstruct the bare SMPL model at a resolution of $512\times 512$ using the hyper-parameters suggested by~\cite{zhao2022human} for $40,000$ steps, which takes approximately $20$ minutes. For avatar creation, the coarse training involves $40$ epochs with $100$ body captures and $20$ head captures for each epoch, while the fine training involves $10$ epochs with $100$ body captures and $50$ head captures. We train both stages using the Adam~\cite{kingma2014adam} optimizer with a learning rate of $5e-3$. The coarse stage takes around $30$ minutes, while the fine stage takes $100$ minutes. For Score Distillation, we use pretrained Stable Diffusion~\cite{rombach2022high} v1.5 model, with a guidance scale of $100$ for all prompts. We conduct experiments on one NVIDIA A100 GPU.

\subsection{Qualitative Result}

% We report qualitative results in Fig.~\ref{fig:shape} and Fig.~\ref{fig: result}, where a variety of prompts from various styles are tested, including both specific identities and general descriptions.
\final{In this section, we provide qualitative results of AvatarCraft. We will show that the proposed method is effective for appearance generation, geometry generation, concept mixing, as well as parametric articulation.}

\final{\textbf{Appearance Generation.}} As shown in the Fig.~\ref{fig: result}-a, AvatarCraft \final{faithfully generates} the distinctive features of representative identities, such as the muscular body of the character \textit{``Hulk''}, the nose of \textit{``Flynn Rider''}, and the costume of \textit{``Captain America''}. This demonstrates the ability of AvatarCraft to produce textures that are semantically consistent and aligned with the target text descriptions. Additionally, AvatarCraft is capable of generating high-fidelity details not only on the human body but also on the head. For example the machine crew of \textit{``Robot''} and the red cape and black beard of \textit{``Doctor Strange''}.

\final{{\textbf{Geometry Generation.} } In addition to appearance generation, AvatarCraft is also capable of carving geometry details on the avatar body. For instance, in Fig.~\ref{fig:geo_edit} we show the rendered depth map of original bare SMPL and avatar \textit{``Hulk''}. We observe clear muscle structure being formed that is consistent with the generated texture.}

\final{{\textbf{Concept Mixing.} A strength of text-to-image models is their creativity in generating images that mix different concepts together using text guidance. Thanks to the diffusion constraints, our AvatarCraft can generate novel avatars (\ie to dream avatars \cite{hong2022avatarclip,poole2022dreamfusion}) by mixing different identities together. This would allow more fine-grained control over avatar style, clothing, and face. We provide examples for those cases in Fig.~\ref{fig:mix}. As the result shows, our proposed method successfully generates avatars that match the input prompt.}}

\final{\textbf{Parametric Articulation.}} \final{As an articulated-NeRF based method, AvatarCraft} can render novel views with parametric control over poses and shapes simultaneously\final{, while preserving render quality. This is achieved by defining} an explicit warping field, represented by a local transformation between the source and target SMPL meshes.
\final{By referring to the generated avatar in canonical space, this warping operation enables rendering of avatar from novel views,  poses, and shapes, without the need of re-training.} \final{We show shape and pose articulation in Fig.~\ref{fig:shape} and Fig.~\ref{fig: result}-b, respectively.}

\begin{figure*}
    \centering
    \includegraphics[width=0.95\linewidth]{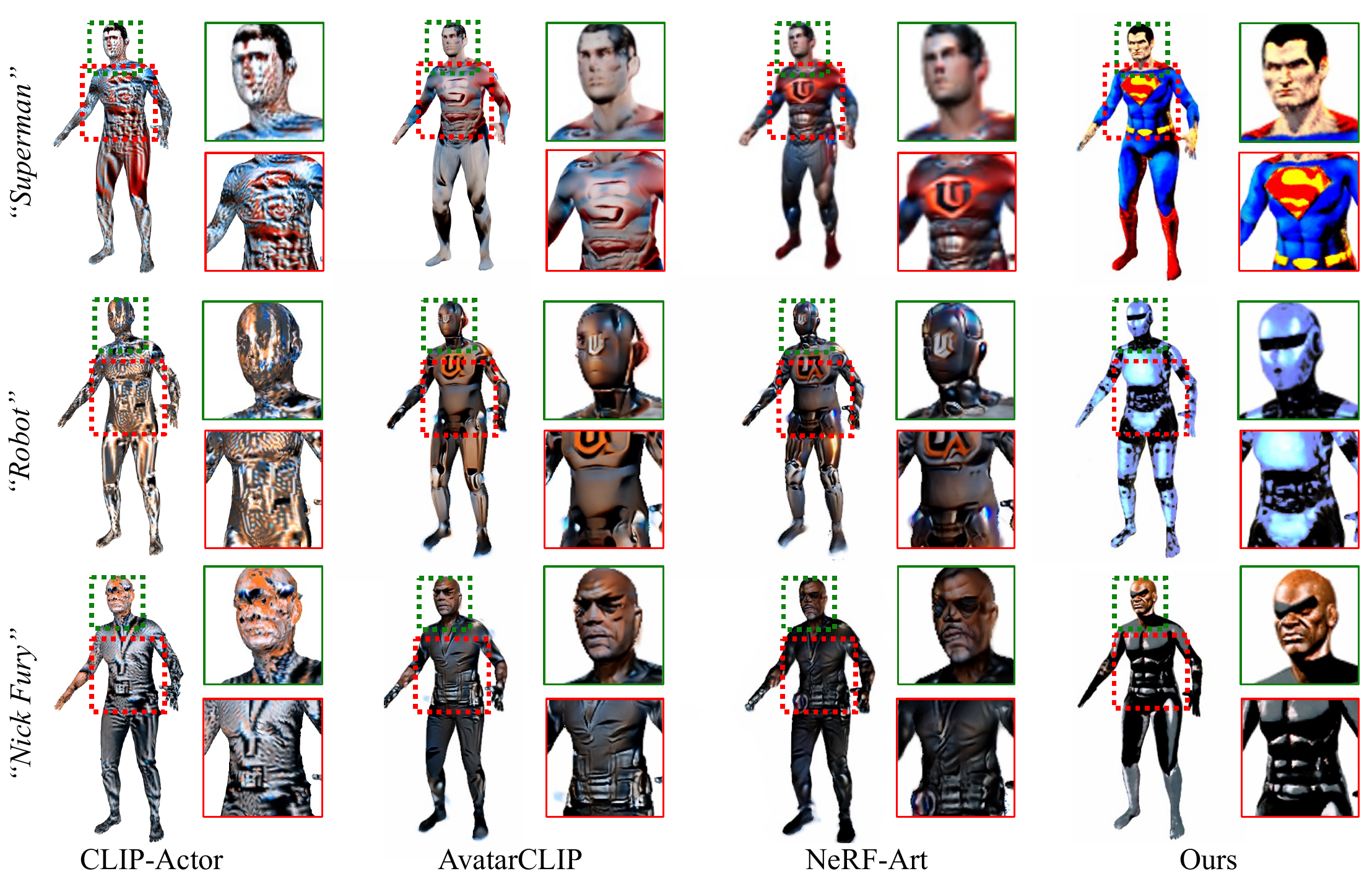}
    \caption{\textbf{Comparisons}. We compare AvatarCraft with state-of-the-art avatar creation methods, including CLIP-Actor~\cite{youwang2022clip}, AvatarCLIP~\cite{hong2022avatarclip}, and NeRF-Art~\cite{wang2022nerf}. Our model achieves better results both globally and locally.}
    % \vspace{-5pt}
    \label{fig: compare}
\end{figure*}

\subsection{Comparisons}

We compare AvatarCraft with state-of-the-art methods for text-driven human avatar creation, including the mesh based method of CLIP-Actor~\cite{youwang2022clip} and \final{implicit-field based} methods of AvatarCLIP~\cite{hong2022avatarclip} and NeRF-Art~\cite{wang2022nerf}. We train each method using the configurations suggested by its respective authors and present results in Fig.~\ref{fig: compare}.

Compared to existing methods, AvatarCraft stands out for the high level of detail in both the \final{avatar} body and head, attributed to its diffusion constraint and our coarse-to-fine and multi-bbox training strategy. For example, it faithfully generates intricate details like the distinctive "S" pattern on \textit{``Superman''}'s chest that other works struggle to capture with equal quality. This demonstrates the strength of AvatarCraft in generating high quality body textures.  In addition, our method produces finely detailed faces, such as the eye patch on \textit{``Nick Fury''}, while other works only present rough facial features.

Another advantage of AvatarCraft is its ability to generate balanced textures for human avatars. This is achieved through the incorporation of a diffusion constraint, which employs a stronger and more advanced language-vision model. In contrast, other works tend to produce unreasonable and uneven textures. \final{We postulate that it may be due to CLIP constraint that perceives and embeds the image as a whole, while being weak in providing location-aware supervision~\cite{zhong2022regionclip,li2023clip} that is necessary for allocating correct texture locally}. \final{In contrast, the diffusion constraint employs a denoising process to provide supervision signals on the pixel level, enabling more localized guidance. Therefore, it could result in more balanced and detailed textures for avatar generation.}

\subsection{Ablation Analysis}
\begin{figure}
    \centering
    \includegraphics[width=0.95\linewidth]{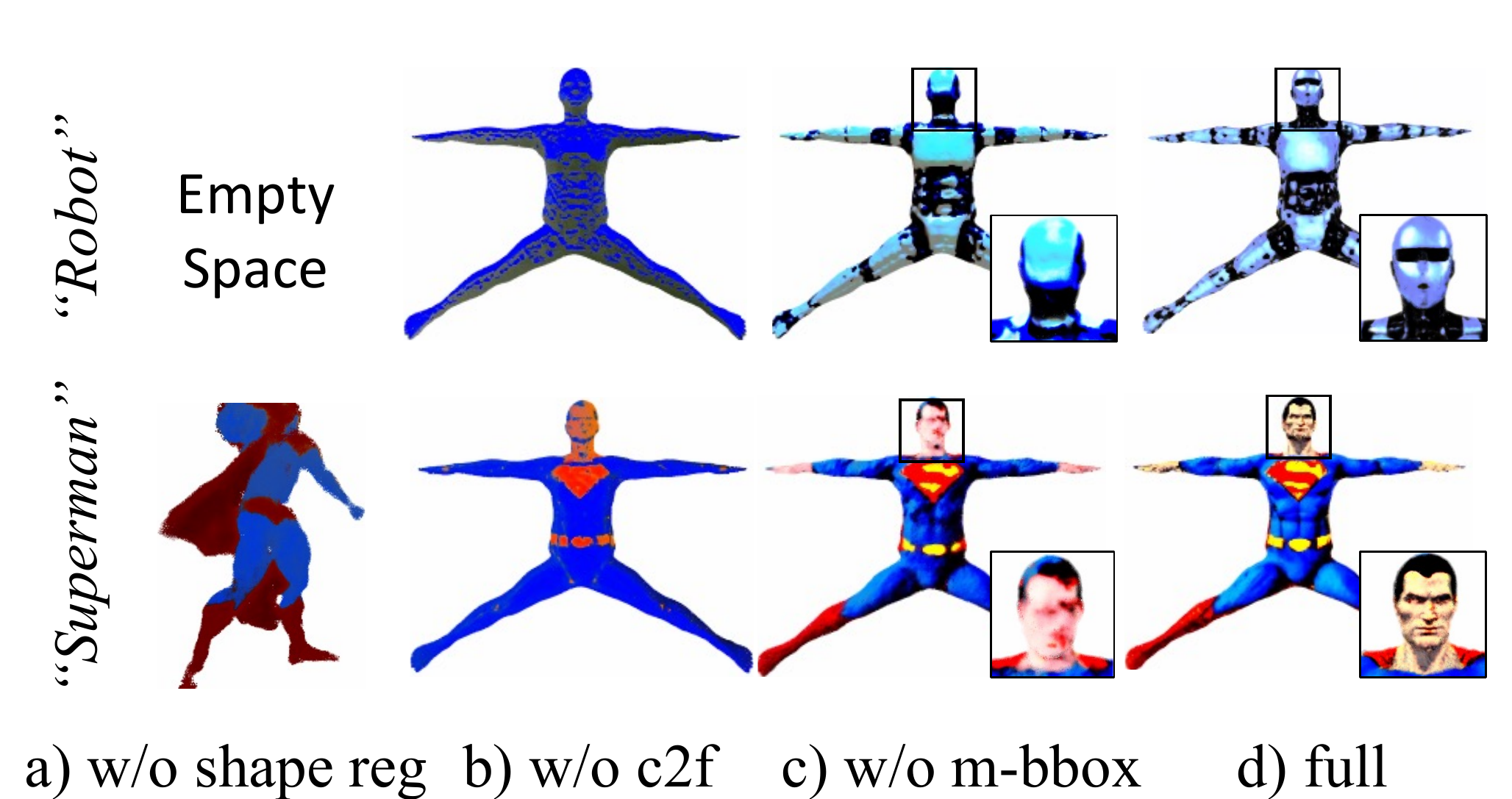}
    \caption{\textbf{Ablation Study.} We show results: (a) without the shape regularization; (b) without coarse-to-fine~(c2f) training; (c) without multi-bbox training; and (d) our full method.}
    % \vspace{-5pt}
    \label{fig: ablate}
\end{figure}

To evaluate the design choices of AvatarCraft, we conduct an ablation study on the effectiveness of 1) the shape regularization, 2) the coarse-to-fine training strategy, and 3) the multi-bbox training strategy. We removed each component from AvatarCraft and reported the results in Fig.~\ref{fig: ablate}. For 2), we train the model at a resolution of $128\times128$ for $20$ epochs. For 3), we train the model only with the body bounding box for the same number of epochs.

Shape regularization is a crucial component of AvatarCraft, as it ensures that the generated avatars have anatomically plausible shapes and proportions. By encouraging the model to learn representations consistent with prior knowledge of human anatomy, shape regularization leads to more accurate avatar creations. Without shape regularization, the method may produce unreasonable or even failed results.

The coarse-to-fine training enables alignment of fine-grained texture generation both globally and locally. In contrast, methods that do not use coarse-to-fine training only produce coarse colors with no semantically meaningful local details.
Our multi-bbox training technique enhances the fidelity of facial features while maintaining a natural transition between the head and body. The methods without multi-bbox training may degrade facial features.

\subsection{Application: Composite Rendering}
\begin{figure}
     \centering
     \includegraphics[width=0.95\linewidth]{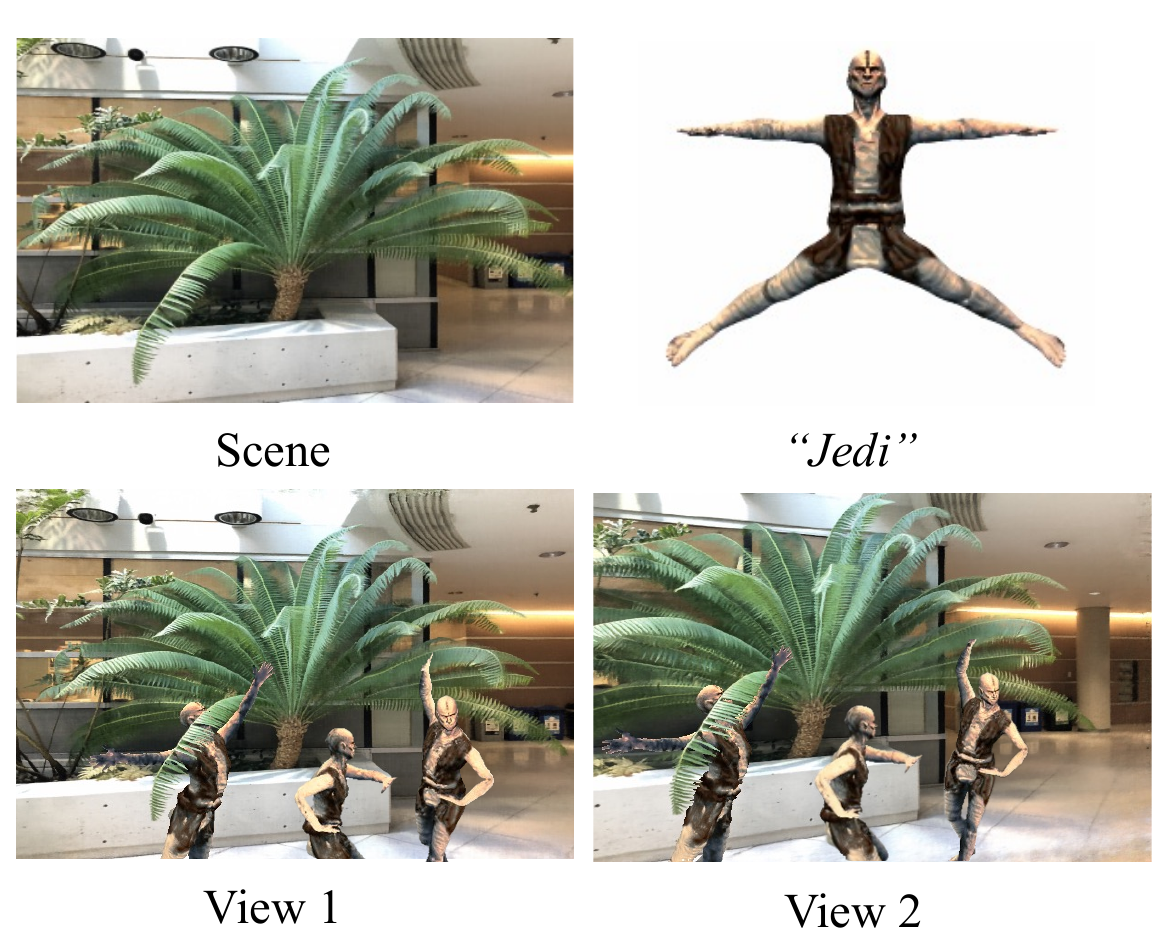}
     \caption{\textbf{Composite Rendering.} AvatarCraft enables occlusion-aware composite renderings of the created avatar with real neural scenes. We show key frames of \textit{``Jedi''} dancing in the scene. }
     % \vspace{-5pt}
     \label{fig:composite}
\end{figure}

Our method enables composite rendering of the avatar along with realistic neural scenes. To achieve this, we manually align the avatar with the scene and render them with the same camera. However, as our avatar is represented as a SDF and implicit scenes are usually encoded as occupancy fields (\eg NeRF), we cannot directly apply Eq.~\ref{eq:render} to a ray that contains points sampled from both representations. To \final{circumvent this challenge}, we primarily perform a depth test to achieve \final{occlusion-aware} composite rendering as illustrated in Fig.~\ref{fig:composite}.

\section{Conclusion}
We present AvatarCraft, a novel approach to generating human avatars using text guidance and an implicit neural representation with parameterized shape and pose control. Unlike existing methods that struggle to produce visually-pleasing results when using CLIP guidance, our approach utilizes diffusion models to match the desired text description and generate fine-grained details in both geometry and texture. Furthermore, our method allows for easy animation and reshaping of the avatar using SMPL parameters and requires no training for novel pose and shape synthesis. Additionally, the use of neural implicit fields provides advantages over mesh-based avatars in terms of photorealistic rendering and easy composition with implicit 3D scenes.

\begin{figure}
    \centering
    \includegraphics[width=0.95\linewidth]{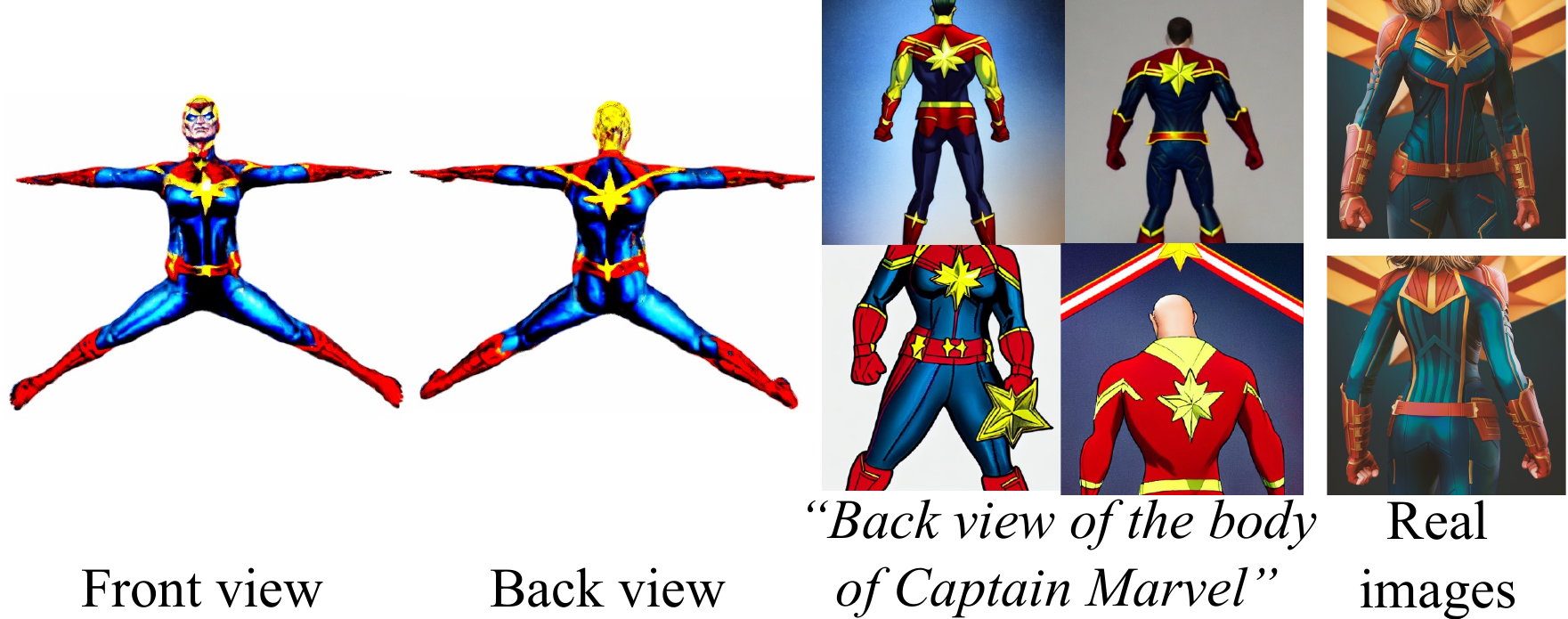}
    \caption{\textbf{Limitation.} We show the created avatar, \final{2D images generated by the base diffusion model, and reference real images.}}
     % \vspace{-5pt}
    \label{fig:limit}
\end{figure}

\textbf{Limitation.} Fig.~\ref{fig:limit} \final{demonstrates one} limitation of our approach, where the diffusion model \final{has difficulty generating textures of equal quality} on the back of the avatar \final{ due to its limited ability to conceptualize views like the back that were underrepresented in the training data}. Consequently, the generated back textures may exhibit certain inconsistencies compared to the \final{reference} ground-truth. To address this issue, we plan to explore the possibility of training the diffusion model with more diverse \final{human} data to enhance its ability to generate faithful and accurate textures.

\section*{Acknoledgement}
\final{This work was supported by a GRF grant (Project No. CityU 11208123) from the Research Grants Council (RGC) of Hong Kong.}

% ========================================
% ========================================

%%%%%%%%% REFERENCES
{\small
\bibliographystyle{ieee_fullname}
\bibliography{main}
}

% ========================================
% ========================================
\clearpage
\appendix

\section{More Implementation Details}

We adopt similar augmentation strategies to those used in recent text-to-3D generative models~\cite{hong2022avatarclip, poole2022dreamfusion,lee2022understanding,jain2022zero,lin2022magic3d}. In this section, we describe the adopted augmentations in detail.

\textbf{Random Camera Extrinsic Augmentation.} During the template avatar creation process, we randomize the camera extrinsic when rendering the implicit avatar $\mathcal{N}$. More specifically, we fix the camera to look at the center and sample elevation angel $\theta \sim \mathcal{U}(-\pi/6,\pi/6)$, azimuth angle $\phi \sim \mathcal{U}((-\pi/3,\pi/3)\cup (2\pi/3, 4\pi/3))$, and camera distance in $\mathcal{U}(2.0,2.2)$.

\textbf{Random Background Augmentation.} We augment the rendering of implicit avatar $\mathcal{N}$ by using random background color. Three types of background are used, including 1) pure white. 2) pure black. 3) Gaussian noise $\mathcal{N}(0.5,0.1)$.

\begin{figure}[!h]
    \centering
    \includegraphics[width=\linewidth]{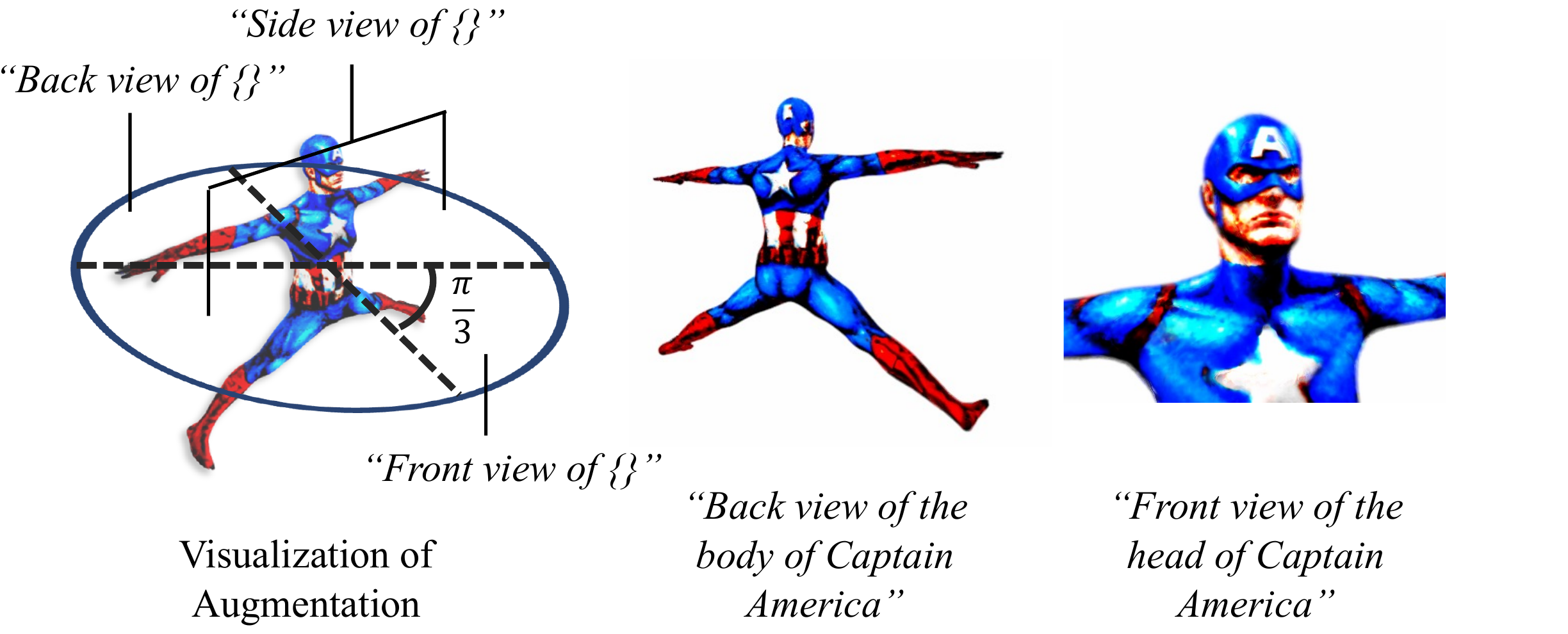}
    \caption{\textbf{Prompt Augmentation.}}
    \label{fig:aug}
\end{figure}

\textbf{View-dependent Prompt Augmentation.} We combine the prompt augmentation technique for human ~\cite{hong2022avatarclip} and for general object ~\cite{poole2022dreamfusion} to provide more semantically meaningful text guidance. First, we use \textit{``the body of \{\}''} and \textit{``the face of \{\}''} when rendering body and head bounding box of avatars, respectively. Additionally, we further augment the prompt by \textit{``front view of \{\}''}, \textit{``side view of \{\}``} depending on the range of azimuth angle $\phi$. More specifically, we use \textit{``Front view of \{\}''} when $5\pi/6\le\phi\le7\pi/6$, and use \textit{``Back view of \{\}''} when $-\pi/6\le\phi\le\pi/6$, otherwise we use \textit{``Side view of \{\}''}. Putting it together, example prompts include "back view of the body of Captain America" and "front view of the head of Captain America". We visualize our augmentation in Fig.~\ref{fig:aug}.

\section{Additional Qualitative Results}
\begin{figure*}
    \centering
    \includegraphics[width=\linewidth]{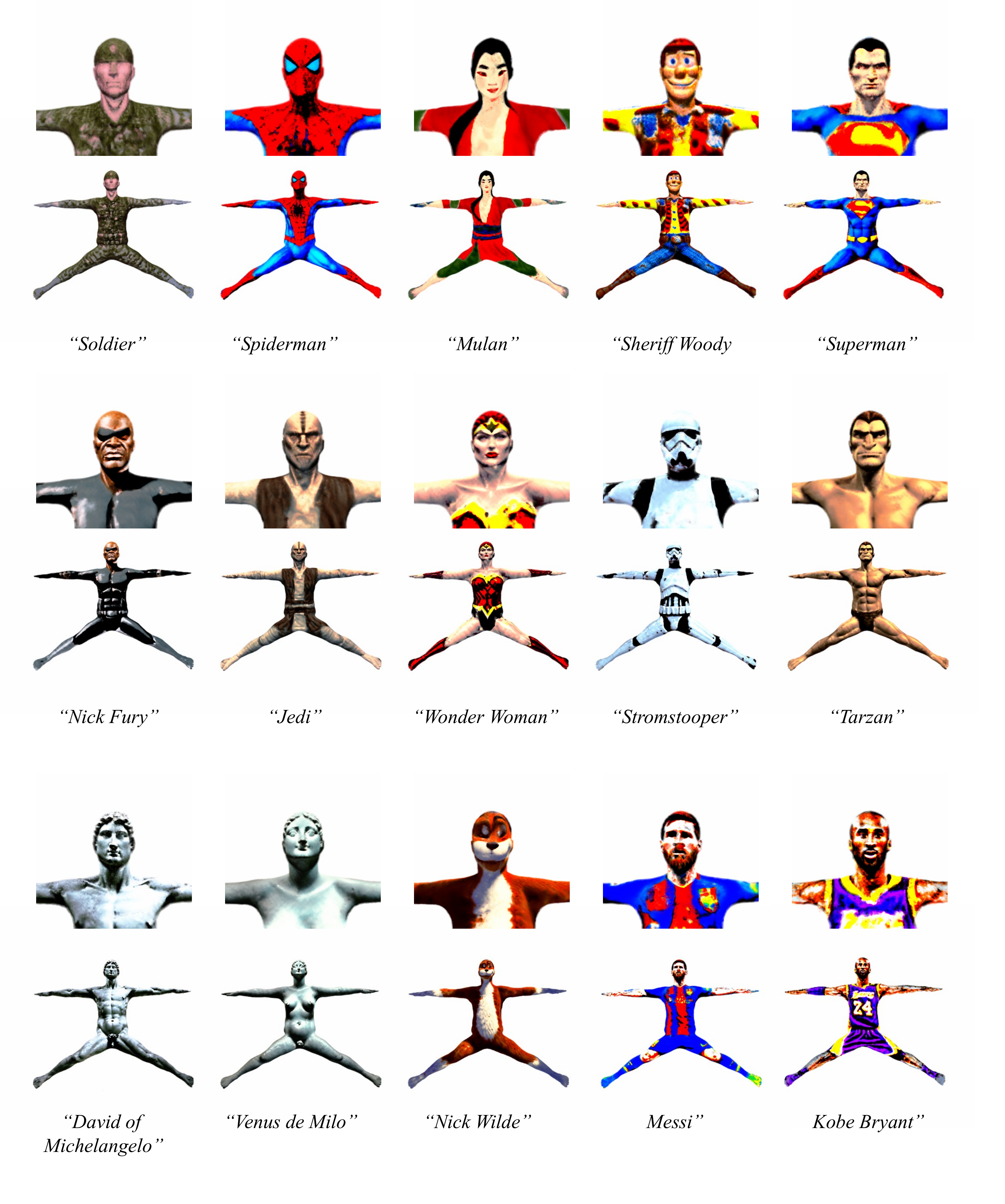}
    \caption{\textbf{More Generated Avatars.}}
    \label{fig:more_1}
\end{figure*}

\begin{figure*}
    \centering
    \includegraphics[width=\linewidth]{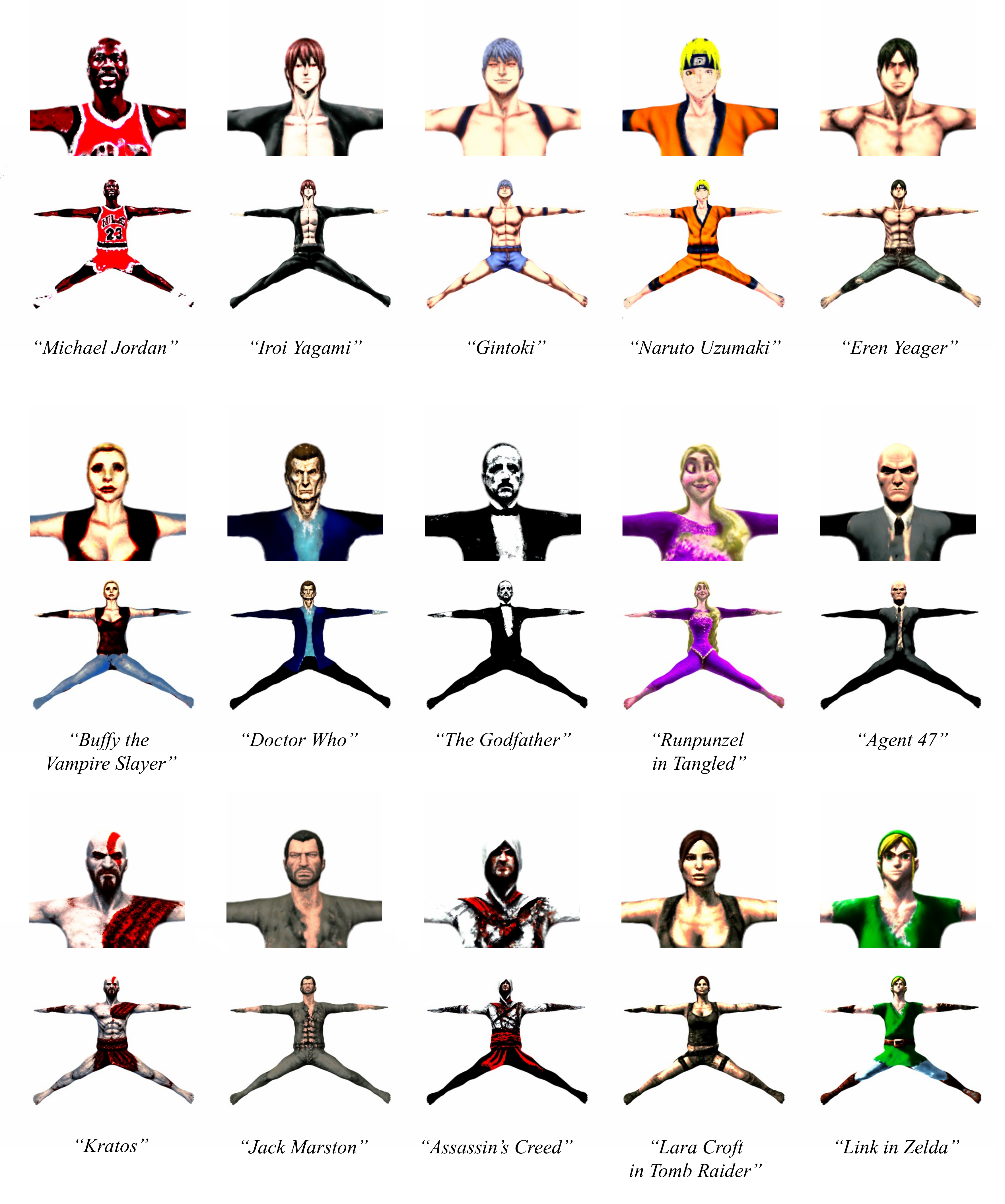}
    \caption{\textbf{More Generated Avatars.}}
    \label{fig:more_2}
\end{figure*}

\begin{figure*}
    \centering
    \includegraphics[width=\linewidth]{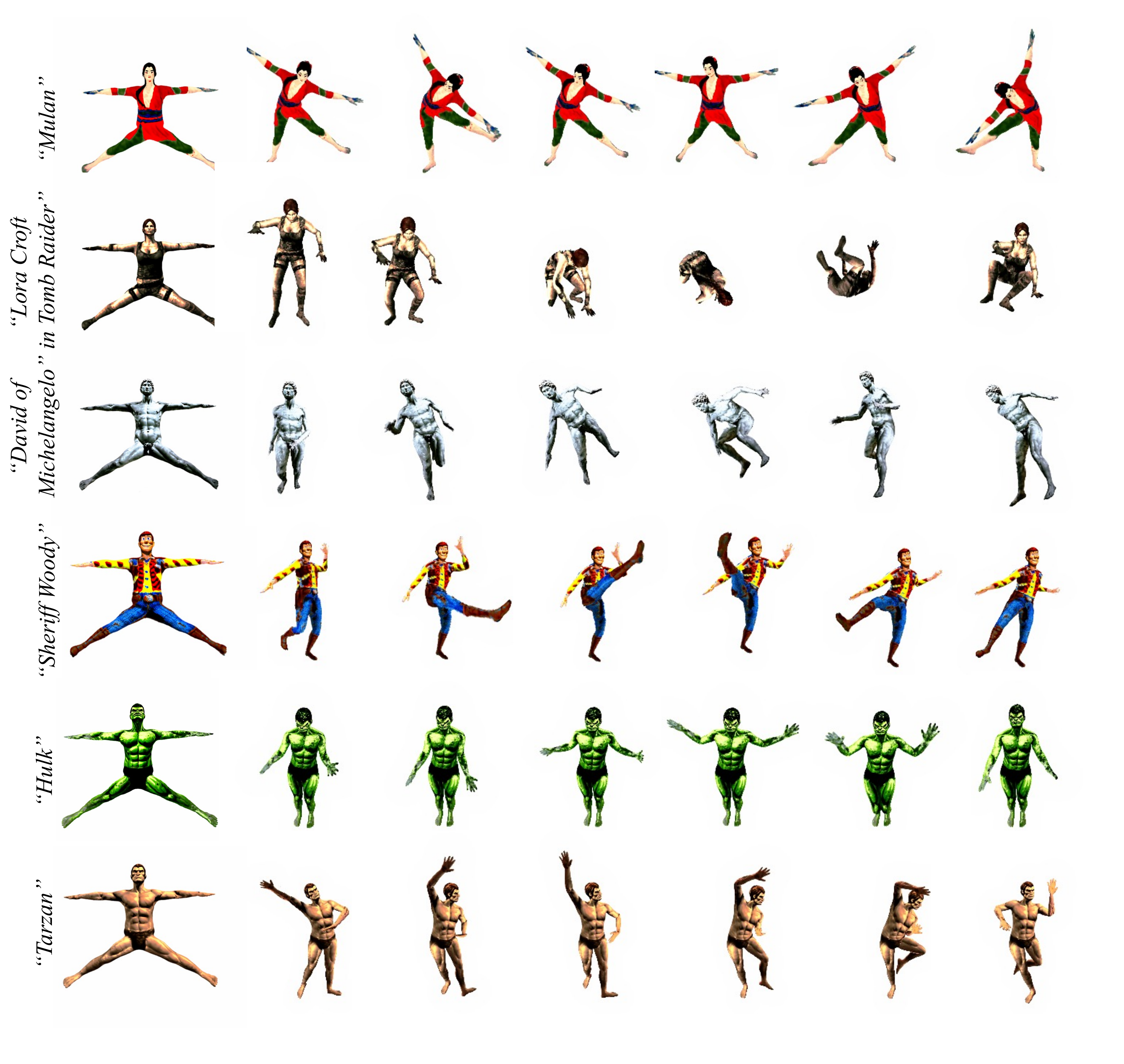}
    \caption{\textbf{Pose Sequences.}}
    \label{fig:pose_seq}
\end{figure*}

In this section, we provide more results of our proposed method, including additional generated avatars in Fig.~\ref{fig:more_1} and Fig.~\ref{fig:more_2}, as well as pose sequences in Fig.~\ref{fig:pose_seq}.

\section{Supplementary Video}
We provide a supplementary video with more visual results rendered in multiple views. We highly recommend watching our supplementary video in \url{https://avatar-craft.github.io/} to observe the user-friendliness and view consistency that our method can achieve in creating, reshaping, and animating neural human avatars.

\end{document}